\documentclass[letterpaper]{degruyter-journal-a}           

\usepackage{algorithm}
\usepackage{algpseudocode}

\pdfoutput=1

\title{Machine Translation in Indian Languages: Challenges and Resolution}

\lastnameone{Patel}
\firstnameone{Raj Nath}
\nameshortone{R.\,N.~Patel}
\addressone{KBCS Division, Centre for Development of Advanced Computing, Gulmohar Cross Road No 9, Opp Juhu shopping Center, Juhu, Mumbai-400049}
\countryone{India}
\emailone{patelrajnath@gmail.com}

\lastnametwo{Pimpale}
\firstnametwo{Prakash B.}
\nameshorttwo{P.\,B.~Pimpale}
\addresstwo{KBCS Division, Centre for Development of Advanced Computing, Gulmohar Cross Road No 9, Opp Juhu shopping Center, Juhu, Mumbai-400049}
\countrytwo{India}
\emailtwo{prakash@cdac.in}

\lastnamethree{M}
\firstnamethree{Sasikumar}
\nameshortthree{M.\,~Sasikumar}
\addressthree{KBCS Division, Centre for Development of Advanced Computing, Gulmohar Cross Road No 9, Opp Juhu shopping Center, Juhu, Mumbai-400049}
\countrythree{India}
\emailthree{sasi@cdac.in}

\newcommand{\enhi}{English-Hindi\xspace}
\newcommand{\enml}{English-Malayalam\xspace}
\newcommand{\enpu}{English-Punjabi\xspace}
\newcommand{\enta}{English-Tamil\xspace}

\newcommand{\en}{English\xspace}
\newcommand{\ta}{Tamil\xspace}
\newcommand{\ml}{Malayalam\xspace}
\newcommand{\hi}{Hindi\xspace}

\abstract{English to Indian language machine translation poses the challenge of structural and morphological divergence. This paper describes English to Indian language statistical machine translation using pre-ordering and suffix separation. The pre-ordering uses rules to transfer the structure of the source sentences prior to training and translation. This syntactic restructuring helps statistical machine translation to tackle the structural divergence and hence better translation quality. The suffix separation is used to tackle the morphological divergence between English and highly agglutinative Indian languages. We demonstrate that the use of pre-ordering and suffix separation helps in improving the quality of English to Indian Language machine translation. 
}



\begin{document}


\section{Introduction}
In this paper, we present our Statistical Machine Translation (SMT) experiments from English to Tamil, Malayalam, Punjabi, and Hindi. From the set of target languages involved, Hindi and Punjabi belong to the Indo-Aryan language family, whereas Tamil and Malayalam belong to the Dravidian language family. All languages except English, have the same flexibility towards word order, canonically following the Subject-Object-Verb (SOV) structure, whereas English follows the Subject-Verb-Object (SVO) structure.

The structural difference between the source and target language makes SMT difficult. It has been demonstrated that pre-ordering benefits SMT in such cases~\cite{ramanathan:2008,rama:2014}. Pre-ordering or reordering transforms the source sentence into a target-like order using syntactic parse tree of the source text. After reordering, training of the SMT system is performed using parallel corpus. Reordering also applies to the new source sentences prior decoding. The reordering system is generally developed using a rich set of rules. These rules are manually extracted based on analysis of parsed source sentence and corresponding target language translation. We used the reordering system developed by~\cite{patel:2013}.

With reference to the morphology, Tamil and Malayalam are more agglutinative compared to English. It is also known that SMT produces more unknown words resulting in bad translation quality, if morphological divergence between the source and the target language is high.~\cite{koehn:2003,popovic:2004,popovic:2006,kumar2014factored} and~\cite{pimpale:2014} have demonstrated ways to handle this issue with morphological segmentation before training the SMT system. To tackle the morphological divergence between English and Tamil we used a suffix separation system developed by~\cite{patel:2014,pimpale:2014} as a pre-processing step. The suffix separation tries to reduce the morphological divergence between the source and agglutinative target language by splitting compound words. Such words when generated in decoding are combined to form a single word using a post-processor.

A factored SMT with the stem as an alignment factor~\cite{koehn:2007:factored} has been trained to achieve better alignment. The target side transliteration is also applied to non translated words.

The rest of the paper is organized as follows. In section~\ref{sec:chalanges}, we discuss the challenges of \en to Indian language machine translation. Section~\ref{sec:setup} describes the dataset and experimental setup. Section~\ref{sec:results} discusses experiments and results followed by a description of our submission to the shared task in section~\ref{sec:sub}. In section~\ref{sec:analysis}, we report few early observations with conclusion and future work in section~\ref{sec:conc}.


\section{Challenges of English to Indian Language SMT} \label{sec:chalanges}
As discussed briefly in the introduction, English to Indian language MT poses challenges of structural and morphological differences. In the following subsections, we discuss the syntactic and morphological divergence with examples.  

\subsection{Syntactic Divergence} \label{sec:order}
The important structural difference in English and most of the Indian languages is the word order. English uses the Subject-Verb-Object (SVO) order and most of the Indian languages, including the ones under study, primarily use Subject-Object-Verb (SOV). Some of the Indian languages are of the nature of free word order. Prepositions in case of \en come after the pronoun or noun they qualify and for \hi they succeed the noun or pronouns, also known as postpositions. Two representative examples are given in table~\ref{tab:exro}. In the first examples, we can see that word order "ate mango" becomes "mango ate" ({\it aama khaayaa}\footnotemark \footnotetext{All \textbf{\textit{Non-English} (Hindi, Tamil)} words have been written in Itrans using~\url{http://sanskritlibrary.org/transcodeText.html}; For Tamil, we have written the word pronunciation in Devanagari and then trans-coded in Itrans}) in Hindi. In the second example, the preposition 'on' (\textit{para} in Hindi) becomes the postposition of the noun phrase 'the table' (\textit{tebala}). 

\begin{table*}[!hbt]
	\centering
	\begin{tabular}{l|l}
		\en Sentence & \hi Sentence \\ \hline
		Ram ate mango & \textit{raama ne} (Ram) \textit{aama} (mango) \textit{khaayaa} (ate)   \\
		Apple is on the table  & \textit{seba} (Apple) \textit{tebala} (the table) \textit{para} (on) \textit{hai} (is)  
	\end{tabular}
	\caption{Example of different word order in \en and \hi}
	\label{tab:exro}
\end{table*}

\subsection{Morphological Divergence} \label{sec:morphy}
We discuss here morphological divergence of Tamil and Malayalam with respect to English using analysis based on the parallel corpus detailed in table~\ref{tab:analysis}. Purpose behind comparing Tamil and Malayalam with English was to demonstrate the difference of agglutination. The morphological divergence for Hindi and Punjabi with respect to English would not be as high as these languages. In our old studies~\cite{pimpale:2014,patel2016statistical} we have compared Marathi, Tamil, Telugu, and Bengali with Hindi, where all the languages were more agglutinative as compared to Hindi.

We know that the parallel corpus represents the same information in two different languages. In table~\ref{tab:analysis}, we can see that \en makes use of more words to represent the same concept or information as compared to \ta and \ml. If we look at the unique words for each language, we can conclude that \en has much less vocabulary as compared to these two Indian languages. This implies that \en needs to make use of different combinations of available words to represent various concepts. Whereas, in \ta and \ml, different concepts are represented by different words. Examples of the same can be seen in table~\ref{tab:exmorph}. The average sentence length of these languages also establishes the same fact. The significant difference in {\it average word length}\footnotemark \footnotetext{Average word length calculated on unique words, on total words, it is 4.8 for \en and stays almost same for others.} shows that the words of \ta and \ml are longer as compared to that of \en. Many new words in these Indian languages are formed by compounding of words or suffixes. The phenomenon is called agglutination and so we say that \ta and \ml are more agglutinative compared to \en.

The difference in length of source and target sentence makes the word alignment difficult. The wrong alignments ultimately results into poor quality translation system. In our experiments, we try to tackle this issue by using suffix separation methods for \enta SMT.

\begin{table*} [!hbt] \small
	\centering
	\begin{tabular}{l|cc|cc} 
	\multicolumn{1}{l|}{} & \multicolumn{2}{c|}{\enml} & \multicolumn{2}{c}{\enta} \\
	\multicolumn{1}{l|}{} & \multicolumn{1}{c}{\en} & \multicolumn{1}{c|}{\ml} & \multicolumn{1}{c}{\en} & \multicolumn{1}{c}{\ta} \\ \hline
	\#sentences  & 103K & 103K & 139K & 139K\\
	\#total words & 1673K & 1069K & 2189K & 1576K \\
	\#unique words & 51K & 209K & 71K & 255K \\
	average word length (\#characters)  & 8.02 & 12.40 & 8.12 & 11.95 \\
	average sentence length (\#words) & 16.31 & 10.42 & 15.75 & 11.33 \\
	\end{tabular}
\caption{Statistical analysis of morphological divergence}
\label{tab:analysis}
\end{table*}

\begin{table*}[!hbt]
	\centering
	\begin{tabular}{l|l}
\en & \ta \\ \hline
have to go & \textit{pokanuma}   \\
that too   & \textit{aTavuma}
\end{tabular}
	\caption{Example \en phrases and equivalent \ta words}
	\label{tab:exmorph}
\end{table*}


\section{System Setup} \label{sec:setup}
In the following subsections, we describe Data distribution followed by pre-processing, evaluation metrics, and SMT system setup used for the experiments.

\subsection{Data set} \label{sec:data}
For our experiments, we used the corpus shared by MTIL-2017~\cite{premjith2016fast}, detailed in Table \ref{tab:data}. We split the shared data into train, test, and development sets. We used  publicly  available {\it Indian language tokenizer and text normalizer}\footnotemark \footnotetext{\url{http://anoopkunchukuttan.github.io/indic_nlp_library/}} for all the target languages. For English, we used tokenizer available with $moses$\footnotemark \footnotetext{\url{https://github.com/moses-smt/mosesdecoder}}. For long sentences, Expectation Maximization (EM) algorithm has hard time to learn the word alignments. Also if the source-to-target word length ratio is very high implies misaligned segments. So, we removed the sentences having word count > 80 or source-target word length ratio > 1:9. 

\begin{table*}[hbt]
	\centering
	\begin{tabular}{l|cc|cc|cc}
		\multicolumn{1}{c|}{} & \multicolumn{2}{c|}{training} & \multicolumn{2}{c|}{development} & \multicolumn{2}{c}{testing} \\
		\multicolumn{1}{l|}{} & \multicolumn{1}{l}{\#sents} & \multicolumn{1}{l|}{\#words} & \multicolumn{1}{l}{\#sents} & \multicolumn{1}{l|}{\#words} & \multicolumn{1}{l}{\#sents} & \multicolumn{1}{l}{\#words} \\ \hline
		\multicolumn{1}{l|}{\enml} & 101K & 1846K & 500 & 9134 & 500 & 9450 \\ 
		\multicolumn{1}{l|}{\enhi} & 159K & 2954K & 500 & 8891 & 500 & 9168 \\ 
		\multicolumn{1}{l|}{\enpu} & 128K & 2089K & 500 & 8247 & 500 & 8337 \\ 
		\multicolumn{1}{l|}{\enta} & 138K & 2442K & 500 & 8833 & 500 & 8475 \\ 
		
	\end{tabular}
	\caption{Data distribution}
	\label{tab:data}
\end{table*}

\subsection{Preprocessing} \label{sec:prep}
SMT works well if the structural divergence between source and target language is not very high. To reduce the structural divergence between source and target language, we used source side reordering. To tackle the morphological divergence between the source and target, we preprocessed the Tamil with suffix separation.

\paragraph{Reordering (RO)} is a preprocessing stage for Statistical Machine Translation (SMT) system where the words of the source sentence are restructured as per the syntax of the target language prior training. The test set is also preprocessed similar to the training data prior decoding. The idea is to facilitate the training process by better alignments and parallel phrase extraction for a phrase-based SMT system. Reordering also helps the decoding process and hence improving the machine translation quality. A detailed analysis of reordering, improving the training and translation quality is done in~\cite{gupta:2012}.

For English-Hindi SMT, earlier reordering is used by~\cite{ramanathan:2008,rama:2014,patel:2013} and have shown significant improvements over baseline.~\cite{kunchukuttan:2014} reported SMT results for English to {\it 10 major Indian languages}\footnotemark \footnotetext{Hindi, Urdu, Punjabi, Bengali, Gujarati, Marathi, Konkani, Tamil, Telugu, and Malayalam.} and showed that reordering helps for all of them. 

Other language pairs have also shown significant improvement when reordering is employed.~\cite{xia:2004} and~\cite{wang:2007} have observed improvement for French-English and Chinese-English language pairs respectively. \cite{niessen:2004} have proposed sentence restructuring whereas~\cite{collins:2005} have proposed clause restructuring to improve German-English SMT.~\cite{popovic1:2006,popovic2:2006} have also reported the use of simple local transformation rules for Spanish-English and Serbian-English translation. Recently, \cite{khalilov:2011} proposed a reordering technique using a deterministic approach for long distance reordering and non-deterministic approach for short distance reordering exploiting morphological information. Some reordering approaches are also presented exploiting the SMT itself~\cite{gupta:2012,dlougach:2013}. 

\paragraph{Suffix Separation (SS)} is the process where the words are split into stem and suffixes. For machine translation, the splitting of an unknown word into its parts enables the translation of the word by the translation of its parts. For example (Hindi-Marathi SMT), in Marathi, '\textbf{\textit{mahinyaaMnii}}' is translated as '\textbf{\textit{mahiine meM}}' (in the month) in Hindi. In this case, we split the word into '\textbf{\textit{mahiny + aaMnii}}'. Here, the suffix '\textbf{\textit{aaMnii}}' corresponds to the word '\textbf{\textit{meM}}' in Hindi.

We considered only suffixes from target language (Tamil) which corresponds to preposition in the source language (English). For this task, the list of suffixes (\#suffixes = 16) is manually created with the linguistic expertise. When a word is subjected to SS, the longest matching suffix from the list is considered for the suffix separation. The suffix separation takes place only once for a word. We add a $continuation$ symbol "$@@$" after the stem word (mahiny@@), which is used to combine the suffixes back after translation. Pseudo-code for the suffix separation is detailed in Algorithms~\ref{algo:ss}.

\begin{algorithm*} \small
	\caption{Suffix Separation}\label{algo:ss}
	\begin{algorithmic}[1]
		\Procedure{SuffixSep}{$word$}
		\State $\textit{suffixSet} \gets \text{read file } \textit{suffix list}$
		\State $\textit{splits} \gets \textit{\{word, "NULL"\}}$
		\For{$\text{suffix} \gets \text{suffixSet}$}
		\If $\textit{ word.ENDSWITH} = \text{suffix} \And \textit{word.LENGTH} > \textit{suffix.LENGTH}$
		\State $\text{splits[0]} \gets \textit{word.SUBSTRING(0, word.LASTINDEXOF(suffix))} + "@@"$
		\State $\text{splits[1]} \gets \textit{suffix}$
		\Return $splits$
		\EndIf
		\EndFor
		\EndProcedure
	\end{algorithmic}
\end{algorithm*}

Many researchers have tried compound word splitting and suffix separation for SMT between morphologically rich languages.~\cite{brown:2002} has proposed an approach guided by a parallel corpus. The work is limited to breaking compounds into cognates and words found in a translation lexicon, but no results on translation performance are reported.~\cite{koehn:2003:CS} have demonstrated an empirical method of learning the compound splitting using monolingual and bilingual data and reported impact on performance of SMT.~\cite{patel:2014,pimpale:2014} reported significantly improved translation quality for Indian languages SMT using suffix separation and compound word splitting.  

\subsection{Transliteration}
Out-of-Vocabulary (OOV) words occur in almost all Machine Translation (MT) systems. These words are mostly named entities, technical terms or foreign words that were not part of training corpus or were not added to the development dictionary. So, OOV words need to be translated to the target language using transliteration. Transliteration helps to improve the translation quality~\cite{patel:2014} and it has also been shown to be useful for translating closely related language pairs~\cite{durrani:2010,nakov:2012}. For most of the language pairs parallel corpus of transliterations isn't readily available and even if such a training data is made available, the arrangement to integrate transliterated words into MT pipelines are not available in SMT toolkits like phrasal~\cite{green:2014} and joshua~\cite{post:2015}.

Generally, a transliteration system is trained separately outside of an MT pipeline using supervised training methods. It gives all possible target transliterations for a given source word. Generally, the 1-best output is selected as transliteration and is used to replace the OOV word in the translation, post decoding.

This paper uses unsupervised model~\cite{durrani:2014} based on the Expectation Maximization (EM) to induce transliteration corpus using parallel data, which is then used to train a transliteration model. The implementation is available with the $moses$\footnotemark \footnotetext{\url{https://github.com/moses-smt/mosesdecoder}} toolkit. We use top 100-best transliteration output for OOV words. These candidates are plugged in the translation replacing OOV words and re-scored with the language model to get the best translation for source sentence.

\subsection{SMT system set up}
The baseline system was setup by using the phrase-based model~\cite{och:2003,brown:1990,marcu:2002,koehn:2003} and~\cite{koehn:2007:factored} was used for factored model. We tuned the model parameters using minimum error rate training (MERT)~\cite{och:2003}. The language model was trained using KenLM~\cite{heafield:2011} toolkit with modified Kneser-Ney smoothing~\cite{chen:1996}.We tried various n-gram language models and found that 5-gram performs best for the languages under study. For factored SMT training source and target side stem has been used as alignment factor. Stemming for Hindi, Punjabi, Tamil, and Malayalam, has been done using a modified version of lightweight stemmer~\cite{ramanathan:2003:stem}. For English we have used porter stemmer~\cite{minnen:2001}.

\subsection{Evaluation metrics}
The different experimental systems are being compared using, BLEU~\cite{papineni:2002}, PER~\cite{popovic:2007}, TER~\cite{snover:2006:ter}, and CDER~\cite{leusch:2006}. For an MT system to be better, higher BLEU scores with lower PER, TER, and CDER are desired.


\section{Experiments and Results} \label{sec:results}
We carried out various experiments to achieve better accuracy, using the data and system setup described in previous sections. Table~\ref{tab:results} details the experiments we tried. We report BLEU, 1-TER, 1-PER, and 1-CDER for the various experiments. TER, PER and CDER are the word error rates (WER) to measure the quality of translation. In general, these scores should be low for a better MT system. In contrast, 1-WER implies that higher is the value better would be the accuracy. It can be seen that the use of preprocessing and transliteration has contributed to the improvement of 1 to 1.5 BLEU points over the baseline for \enhi, \enpu and \enta. For \enml the BLEU has decreased and we plan to investigate this in our future work. Also, investigation is needed to figure out why the BLEU score decreased on use of factors in \enpu, while it was useful for other language pairs.

Table~\ref{tab:example} describes with an example how reordering reduces the structural divergence and helps to achieve better translation quality. From the example, it can be seen that the translation of the system using S3 is better than S1. The output of S3 is structurally more correct and conveys the same meaning as that of the reference translation.

\begin{table}[!hbt]
	\centering
	\begin{tabular}{l|c|c|c|c|c}
			  & & BLEU &  1-TER & 1-PER & 1-CDER \\ \hline
		\enml & S1 & 08.52 & 13.63 & 32.32 & 21.46 \\
			  & S2 & 08.15 & \textbf{14.37} & \textbf{32.74} & \textbf{21.57} \\
			  & S3 & 08.10 & 09.85 & 24.07 & 20.36 \\
			  & S4 & \textbf{08.25} & 10.03 & 24.38 & 20.52 \\ \hline
		\enhi & S1 & 16.75 & 27.05 & 51.73 & 33.95 \\
		      & S2 & 18.74 & 31.30 & 51.94 & 37.37 \\
		      & S3 & 19.30 & 33.38 & 52.35 & 37.61 \\
		      & S4 & \textbf{19.43} & \textbf{33.53} & \textbf{52.57} & \textbf{37.77} \\ \hline
		\enpu & S1 & 21.71 & 38.26 & 56.13 & 41.44 \\
			  & S2 & \textbf{23.09} & \textbf{40.90} & \textbf{56.83} & \textbf{44.06} \\
			  & S3 & 22.17 & 39.20 & 56.25 & 42.77 \\
			  & S4 & 22.26 & 39.35 & 56.48 & 42.88 \\ \hline
		\enta & S1 & 06.20 & 13.05 & 32.72 & 21.97 \\
			  & S2 & 07.44 & 16.35 & 32.29 & 24.43 \\
			  & $S3\prime$ & 07.47 & 17.87 & 34.86 & 23.49 \\
			  & $S4\prime$ & \textbf{07.56} & \textbf{18.01} & \textbf{35.06} & \textbf{23.62} \\
	\end{tabular}
	\caption{Translation quality scores for different systems; S1: BL; S2: BL+RO; S3: BL+RO+FACT; $S3\prime$: BL+RO+SPLIT+FACT; S4: BL+RO+FACT+TR; $S4\prime$: BL+RO+SPLIT+FACT+TR; BL: Baseline; RO: Reordering; FACT: Factored models; TR: Transliteration}
	\label{tab:results}
\end{table}

\begin{table*}[!hbt] \small
	\centering
	\begin{tabular}{p{1.7cm}|p{11.0cm}}
		Source & Ahmedabad was named after the sultan Ahmed Shah, who built the city in 1411. (\en) \\ \hline
		S1 & ahmedabad was named after the sultan ahmed shah, who built the city in 1411. (\en) \\
		& \textit{ahamadaabaada ke naama para rakhaa gayaa sultaana ahamada shaaha vaale shahara 1411} (machine translated \hi) \\ \hline
		S3 & ahmedabad the sultan ahmed shah after named was , who 1411 in the city built. (reordered \en) \\
		& \textit{ahamadaabaada kaa naama sultaana ahamadashaaha ke naama se paDaa thaa jisane 1411 meM shahara banavaayaa thaa.} (machine translated \hi) \\ \hline
		Reference & \textit{ahamadaabaada kaa naama sultaana ahamadashaaha ke naama para paDaa thaa , jisane 1411 meM shahara banavaayaa thaa.} (manually translated \hi) 
	\end{tabular}
	\caption{Comparison of translation with an example of English-Hindi SMT}
	\label{tab:example}
\end{table*}


\section{Submission to the Shared Task} \label{sec:sub}
As shown in table~\ref{tab:sub-results}, we (C-DACM) submitted our systems for all the language pairs in the shared task. The submitted translations, of the unseen test set, were obtained using S4 and S4$\prime$. The submitted systems were manually evaluated by three native speakers for Adequacy, Fluency, and Rating. The average of the three manual evaluators is given in the Table~\ref{sec:sub}. The shared task organizers (MTIL17) used the percentage of Adequacy and Fluency as the primary metric for the shared task. Evaluation results for the top three participating systems were published by MTIL17~\cite{anand:2018} as shown in the table~\ref{tab:sub-results}. From the evaluation results, it is evident that our (CDAC-M) submissions significantly outperform the other submissions for \enhi, \enta, and \enml. For \enpu our stands the second position. 

\begin{table*}[hbt] \tiny
	\centering
	\begin{tabular}{l|l|c|c|c|c|c|c}
		\multicolumn{1}{l|}{Languages} & \multicolumn{1}{l|}{Team} & \multicolumn{1}{c|}{Avg Adequacy (A)} & \multicolumn{1}{c|}{Avg Fluency (F)} & \multicolumn{1}{c|}{Avg Rating (R)} & \multicolumn{1}{l|}{A\&F\%} & \multicolumn{1}{l|}{R\%} & \multicolumn{1}{l}{BLEU} \\ \hline
		\multicolumn{1}{l|}{\enml} & CDAC-M & 1.92 & 1.67 & 1.60 & {\bf 38.34} & {\bf 31.94} & 2.60 \\ \hline
		\multicolumn{1}{l|}{\enhi} & CDAC-M & 3.82 & 3.63 & 3.43 & {\bf 74.53} & {\bf 68.64} & 20.64 \\ 
		& NIT-M & 3.27 & 3.56 & 3.26 & 68.27 & 65.14 & 23.25 \\
		& IIT-B & 2.55 & 3.23 & 2.59 & 57.81 & 51.87 & 21.01 \\ 
		& JU & 1.81 & 1.72 & 1.58 & 35.28 & 31.50 & 3.57 \\ \hline
		\multicolumn{1}{l|}{\enpu} & NIT-M & 3.38 & 3.74 & 3.235 & {\bf 67.55} & {\bf 65.05} & 9.24 \\ 
		& CDAC-M & 3.05 & 3.02 & 2.92 & 60.91 & 58.34 & 8.68 \\ 
		& IIT-B & 2.65 & 2.71 & 2.62 & 52.93 & 52.4 & 11.38 \\ \hline
		\multicolumn{1}{l|}{\enta} & CDAC-M & 2.61 & 2.57 & 2.40 & {\bf 52.26} & {\bf 48.00} & 6.15 \\ 
		& HANS & 2.16 & 2.12 & 2.17 & 43.22 & 43.50 & 1.93 \\
		& NIT-M & 1.59 & 1.65 & 1.58 & 31.72 & 31.74 & 1.31 \\ 
	\end{tabular}
	\caption{Submissions at MTIL2017; CDAC-M: Centre for Development of Advanced Computing, Mumbai, India; IIT-B: Indian Institute of Technology, Bombay, India; NIT-M: National Institute of Technology, Mizoram, India; JU: Jadavpur University, Kolkata, West Bengal, India; HANS: New York University, New York City, NY, United States; Avg: Average of three manual evaluation scores.}
	\label{tab:sub-results}
\end{table*}

\section{Error Analysis}
\label{sec:analysis}
A closer look at the performance of these systems to understand the utility of Reordering and Suffix Separation has been done. We report a few early observations.

\subsection{Reordering Errors}
\label{subsec:analysis:ro}
We have used Reordering system developed by ~\cite{patel:2013}. Table~\ref{tab:roerror} details an example of the reordering error. In the example, the phrase sequence 'very useful for losing fat' is wrongly reordered and that has resulted into a wrong translation. The wrong reordering not just affects the structure of the output, but also badly affects the phrase translation.

\begin{table*}[!hbt]
	\centering
	\begin{tabular}{l|p{8.0cm}}
		English & Certain foods are very useful for losing fat. \\ \hline
		Reordered & Certain foods very useful fat losing for are . \\ \hline
		System Output & \textit{kuCha khaadya padaarthoM ko khone ke lie bahuta upayogii phaiTa hote haiM}\\ \hline
		Expected Reordering & Certain foods fats losing for very useful are .                   \\ \hline
		Expected Output & \textit{kuCha khaadya padaartha vasaa khone ke lie bahuta upayogii hote hai}
	\end{tabular}
	\caption{Reordering Errors}
	\label{tab:roerror}
\end{table*}

\subsection{Bad Split}
\label{subsec:analysis:bs}
The suffix separation system developed by ~\cite{pimpale:2014} is used. For \ta, it has limited list of manually created suffixes and hence it doesn't work for many words. As suffixes are crudely chopped without consideration of validity of remaining part, the errors get introduced. Most of the errors belong to the category where the words get split because they end with a suffix from our list though these were not meant to be processed. This causes sparsity of these genuine terms in the data and leads to a wrong translation of those. For example, a genuine word, say '\textbf{\textit{abcd}}' is getting split into '\textbf{\textit{ab}}' + '\textbf{\textit{cd}}' which is a wrong split. As, '\textbf{\textit{abcd}}' is a proper noun and hence should not have been split.To avoid suffix separation of such words, NNP POS tag was tried, but that was stopping many other valid candidates from pre-processing. A word getting split at wrong position was also one of the major error case. For example, a word with character sequence '\textbf{\textit{pqrstu}}' was getting split into '\textbf{\textit{pqr}}' and '\textbf{\textit{stu}}' instead of '\textbf{\textit{pqrs}}' and '\textbf{\textit{tu}}'. In such cases, both suffixes '\textbf{\textit{stu}}' and '\textbf{\textit{tu}}' are valid and so deciding on when it goes wrong is difficult.


\section{Conclusion and Future Work} \label{sec:conc}
In this paper, we presented various systems for English to Hindi, Malayalam, Punjabi, and Tamil machine translation. Factored SMT with suffix separation and reordering performs better. Transliteration as postprocessing further helps to improve the translation quality. Failure of factored SMT for \enpu and \enml would be another thread of this work to be continued. Further, we plan to work towards improving the preprocessing and post-processing techniques for better translation quality and extend the approach to other Indian languages.

\bibliography{article}

\end{document}